# A maturity framework for data driven maintenance

Chris Rijsdijk[1], M.J.R. van de Wijnckel[1,2], Tiedo Tinga[1,2]

[1]*Netherlands Defence Academy, Faculty of Military Sciences, P.O. Box 10000, 1780CA, Den Helder, The Netherlands*
*c.rijsdijk.01@mindef.nl*

[2]*University of Twente, Faculty of Engineering Technology, P.O. Box 217, 7500AE, Enschede, The Netherlands*
*m.j.r.vandewijnckel@utwente.nl*
*t.tinga@utwente.nl*

**ABSTRACT**

Maintenance decisions range from the simple detection of faults to ultimately predicting future failures and solving the problem. These traditionally human decisions are nowadays increasingly supported by data and the ultimate aim is to make them autonomous. This paper explores the challenges encountered in data driven maintenance, and proposes to consider four aspects in a maturity framework: data / decision maturity, the translation from the real world to data, the computability of decisions (using models) and the causality in the obtained relations. After a discussion of the theoretical concepts involved, the exploration continues by considering a practical fault detection and identification problem. Two approaches, i.e. experience based and model based, are compared and discussed in terms of the four aspects in the maturity framework. It is observed that both approaches yield the same decisions, but still differ in the assignment of causality. This confirms that a maturity assessment not only concerns the type of decision, but should also include the other proposed aspects.

**1. INTRODUCTION**

Von Leibnitz already dreamt of a universe where decision problems were solved by computations rather than by furious debates. Centuries later, it is much better understood that Von Leibnitz's dream cannot come true. So, one may compute many decisions, but not any decision. Where computed engineering decisions fail, maintenance decisions are typically triggered. Unsurprisingly, maintenance decisions are often hard to compute, or they may even be fundamentally incomputable. However, an inability to compute a decision does not imply that such a decision cannot be supported by computations. This paper will present a maturity framework for computational maintenance decision support.

In this framework, maturity grows as more (advanced) decisions in a maintenance control loop are computed. However, the presented framework not only considers the type of decision, as in existing data maturity models, but relates maturity also to: (*i*) the translation of reality to data (vice-versa), (*ii*) the computability (with models) of the decisions involved and (*iii*) the causality of the relations obtained. A case study will be used to explore the attainable maturity starting from the lowest level. An experience based and a model based approach will be attempted, which both will prove to take the correct decision for an arbitrary validation set. Still, decision makers should care about the approach as causality is managed differently. In the experience based approach, causality will be assigned afterwards. In the model based approach, causality is inherent, as a model that is posited as true is solved. Further, it is observed that it is impossible to compute a true model from only a history of measurements. Therefore, a history of measurements will be indecisive about the approach. Still, the engineering profession established a plethora of guidelines that have often proved to be correct. As these engineering guidelines strengthen (a suspicion of) causality for both approaches, the attainable maturity in data driven maintenance may rise at an acceptable risk.

This paper is organized as follows. Section 2 will introduce the four basic aspects of the framework to assess the maturity in data driven maintenance. Section 3 will portray a typical construction of two different autonomous fault detection and isolation methods (the first step in maturity). Section 4 will demonstrate fault detection and isolation in an iconic case study. Finally, section 5 will discuss the results and section 6 will present the conclusion.

**2. BACKGROUND**

This section will introduce the four basic elements that jointly determine the maturity in data driven maintenance and thus constitute the proposed framework. Section 2.1 addresses the challenges in computing a "real" decision, section 2.2 will discuss the challenges in using (engineering) models to compute decisions. Then section 2.3 will relate the flow of the maintenance control loop with a conventional data maturity model. Finally, section 2.4 discusses the difference between observed associations and causality, and its effect on decision making.

**2.1. Obstructions in computing "real" decisions**

Data (Latin: givens) are input symbols to a syntactical formal language. Hilbert dreamt of a formal language that could provide a complete, consistent, and decidable foundation of mathematics. Gödel (1931), Church (1936) and Turing (1937) showed that such a formal language is nonexistent and the dreams of Von Leibnitz and Hilbert were destroyed. This means that some problems are fundamentally incomputable. Moreover, even the most potent computing devices may just fail to compute a problem in time. Therefore, problems that are computable in principle may be too complex to compute in practice.

A formal language becomes meaningful by assigning a truth value. Then, a computation may become similarly meaningful and it may eventually represent some reasoning about truth or falsehood. Evidently, Von Leibnitz similarly hoped to compute meaningful decisions as he hoped to settle legal disputes this way. Then, a computed decision involves both syntax and semantics (Figure 1).

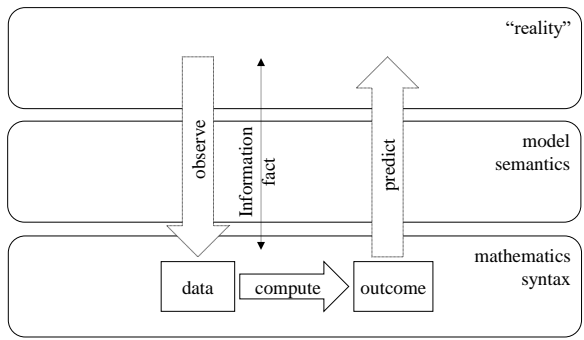

Figure 1. Framework for computing "real" decisions.

Engineers are not necessarily orthodox positivists but the engineering profession is not just about modelling, it also includes building. To circumvent or at least alleviate philosophical controversy, the "reality" in Figure 1 could also be seen as merely a user interface by skeptics who doubt the existence of space-time (Hoffman, 2019).

Any vertical translation in Figure 1 involves an arbitrary human choice, i.e. facts are *made* (Latin: facere) and information is *shaped* (Latin: formare). So, facts and information do not follow from some computable coding operation, they involve arbitrary human choice. For example, observing is not just a mechanical decoding of sound or light waves, it also involves a specific interpretation. Likewise, predicting involves more than just computing an outcome (100101…).

In conclusion, computing a "real" decision may be impossible because (*i*) it is fundamentally incomputable, (*ii*) it is too complex to compute in time, or (*iii*) the translation between "reality" and the syntactical computation is philosophically controversial.

**2.2. Maintenance decisions are incomputable**

A decision (Latin: cut-off) is the elimination of outcomes that would have occurred otherwise. A computation is a deterministic discrete operation that can be performed on a Turing Machine. In a way, a Turing Machine decides as it halts at a particular outcome (while eliminating all other candidate outcomes). So, syntactical decisions include the acceptance or rejection of a string as a well formed formula in a formal language. However, "real" decisions include a choice that causes a specific outcome, rather than any other outcome.

The computation of a "real" decision requires translations between a syntactical Turing Machine and "reality" (Figure 1). These translations are essential for data driven maintenance where computations from syntactical data should support "real" maintenance decisions. Generally, the engineering profession established a high degree of common sense regarding these potentially controversial translations. This common sense has been made explicit in guidelines that specify the computation of the quality of a design (CEN, 2007), (IACS, 2024). Quality is defined by ISO (2015):

> The degree to which a set of inherent characteristics of an object fulfils requirements.

So, quality reflects a margin between measurable inherent characteristics and subjective requirements. So, quality is not just a measurable "real" variable (Figure 1), rather quality is the result of an arbitrary translation between a measurable reality and some subjective aspiration. Engineers showed a great ability to compute outcomes that (often) appeared to satisfy quality in practice. Also in this case, computations from syntactical data support "real" engineering decisions.

The Church-Turing thesis states:

> If something is computable on a discrete device, then it is also computable on a Turing Machine.

This implies that up until now, no one has been able to construct a discrete computing device for which an equivalent Turing Machine does not exist. Still, some computations that are computable on a Turing Machine in principle, may be too complex to compute on a practical device in time. Engineers showed great ability in constructing devices that autonomously compute "real" decisions as feedback control loops are ubiquitous. So, Von Leibnitz's dream *often* became attainable after all. As an example, the feedback controller (C) shown in Figure 2 autonomously computes an input signal (U) to the process (P) that yields an output (Y). This computation depends on the error (W) between the output (Y) and the set point.

Still, the delimitations from section 2.1 remain unresolved implying that (*i*) engineering guidelines are occasionally improved by lessons learned from "real" disasters, or that (*ii*) the feedback control loop occasionally oscillates away from the set point. Where engineering computations fail, maintenance is often triggered. Maintenance is defined by:

> The combination of all technical and administrative actions, including supervision actions, to retain or to restore an item's quality.

This definition paraphrases CEN (2019) and IEC (2015). So, maintenance is considered as a decision to act, with the intention to cause a quality effect. Figure 2 shows a maintenance control loop that should correct the faults of an autonomous feedback control loop (Tinga et al., 2023). The maintenance control loop is typically triggered by the detection of a fault, i.e. an observation of some anomaly. Fault isolation is the assignment of a specific fault label that assists in the choice of the recovery action. Fault identification is an assessment of the (evolution in the) magnitude of the fault. Prognostics is an estimation of the remaining useful life. Finally, recovery is an action that causes quality. This maintenance control loop follows a Fault Detection and Isolation (FDI) convention (Isermann, 2006).

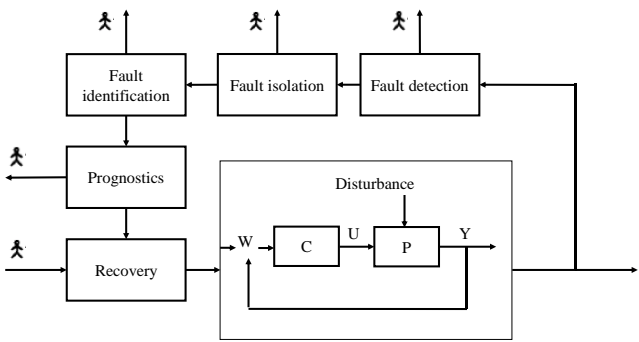

Figure 2. Autonomous control loop extended with maintenance control.

Although the maintenance control loop is thought to be human involved (as indicated by the person symbols in Figure 2), parts of it may still be computed. For example, the fault detection and the fault isolation may be computed before a human takes over. Then, this human may not need to troubleshoot the anomaly as this has been computed autonomously.

In conclusion, engineers have developed a great ability to compute "real" decisions and to construct devices that could similarly do so autonomously. Still, engineering computations occasionally fail which triggers human involved maintenance. Therefore, computing autonomous maintenance is challenging, but parts of the maintenance control loop may still be supported by computations. For that reason, the title of this paper refers to data driven maintenance rather than autonomous maintenance.

### 2.3. Maturity in data driven maintenance

Data maturity models are widely researched (Al-Sai et al., 2023) and applicable. Figure 3 shows a commonly adopted data maturity classification that includes monitoring, understanding, predicting, and deciding.

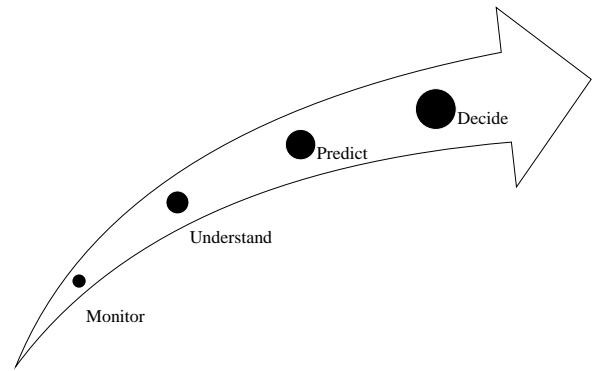

Figure 3. Data maturity model.

A comparison of the data maturity model in Figure 3 with the maintenance control loop in Figure 2 reveals that data maturity grows as more steps in the maintenance control loop are being computed, i.e. monitoring corresponds with fault detection, understanding with fault isolation & identification, predicting with prognosis and deciding with recovery.

Tiddens et al. (2023) observed a relation between an aspired maturity level and the required measurements. This paper intends to be more precise about this relationship by comparing two computations of fault detection and isolation that both provide a correct decision for a specific set of measurements. Still, these two computations will differ in attainable maturity as they translate to "reality" in a different way (Figure 1), i.e. the "real" causal implication of corresponding syntactical computations will be shown to differ. Then, the attainable maturity does not just rely on measurements, but also on a subjective translation.

### 2.4. Causality

This section will introduce two ways to address causality when computing a "real" decision (e.g. in the case study in the next section). In the experience based approach, a statistical association is computed and the causal assumptions are made separately. In the model based approach, the effect of setting a variable in an engineering (design) model of equivalences is computed and the causality follows from the deterministic process of the computation itself.

An equivalence is symmetrical, reflexive, and transitive:

$$Y = aX + b \tag{1}$$

A causality is only transitive:

$$Y \leftarrow aX + b \tag{2}$$

In Eq. (1) swapping the terms around the equivalence symbol does not change the meaning of the expression. However, in Eq. (2) swapping the terms around the arrow changes the meaning of the expression from "$X$ causes $Y$" to "$Y$ causes $X$". It is important to realize that statistical associations retrieved from measurements are equivalences, but they do not imply causality.

Decisions rely on causality rather than on associations as a choice should bring about an effect that would not or less likely occur otherwise. To validate an individual decision, the effect of each choice would have to be observed whereas only the effect of the choice made is observable. Generally, the problem of observing the causal interactions in an individual experiment is that the counterfactuals remain unobservable. Therefore, the interventional distribution $Pr(Y|do(X))$ may wildly differ from the observed distribution $Pr(Y|X)$.

Still, there are ways to strengthen a suspicion of causality across many experiments provided that the cause $X$ in Eq. (2) sufficiently varies. Fisher (1935) proposed random assignment of treatments to eliminate the effect of unobserved confounders and he suggested that unobserved confounders could explain the measured association between smoking and lung cancer (Fisher, 1958). The latter beautifully illustrates the delicacy to use a measured association to support a decision to smoke. Structural Causal Modelling (SCM) proposed by Pearl (2009) also applies to non-experimental research constructs. SCM subsumes Structural Equations Modelling (Wright, 1934), and the Potential Outcomes Framework (Rubin, 2005). The experience based approach to the case study in section 4 will use SCM to specify the independence assumptions needed for a specific causal explanation of a computed statistical association.

Engineers typically use equivalence relations like bond graphs or finite element methods when designing a device. These equivalence relations are acausal, but the computation of their solution is a sequential process that introduces causality, i.e. if one variable in these equations has been set to a known value, the response of the other variables follows by computation. So, there is an intimate relationship between computing the solution of an engineering model and causality (Karnopp et al., 2012). The causal effect of a "real" decision to set one of these variables is similarly computable. The model based approach to the case study will use a bond graph to model the case study and the causality follows from the sequence in the computation itself.

In conclusion, this subsection showed that causality could be assigned after the computation of a statistical association and that causality is just inherent to the process of computing. Both notions of causality will be applied to the case study.

Now these four basic ingredients of data-driven maintenance decision making have been considered, the theoretical concepts will be converted to a practical application in the next two sections.

**3. AUTONOMOUS FAULT DETECTION AND ISOLATION**

This section will portray a typical construction of autonomous fault detection and isolation. Fault detection and isolation are the first "real" decisions in the maintenance control loop (Figure 2). A Fault Signature Matrix (FSM) will be used to assess the ability to detect or isolate faults. The rows in a FSM list the applicable faults (Table 1). A fault can be defined as an anomaly that precedes a failure (= nonconformity in quality). The columns in a FSM list the features (or symptoms) that indicate the faults (Table 1). The fields in a FSM indicate the relationship between the faults and the symptoms. A FSM could therefore support decisions to detect or to isolate faults (step 1 and 2 in Figure 2). For example, $Fault_0$ in Table 1 is detectable and isolable by the feature $F_0$. $Fault_1$ and $Fault_2$ are detectable but not isolable by the features $F_1$ and $F_2$, while $Fault_3$ is both detectable and isolable by these two features.

Table 1: Example of a FSM.

|  | $F_0$ | $F_1$ | $F_2$ |
|---|---|---|---|
| $Fault_0$ | 1 | 0 | 0 |
| $Fault_1$ | 0 | 1 | 1 |
| $Fault_2$ | 0 | 1 | 1 |
| $Fault_3$ | 0 | 0 | 1 |

An Experience Based (EB) and a Model Based (MB) approach to construct a FSM will illustrate two scenarios for the assignment of causality. It will become clear that an EB_FSM merely relates faults to associated symptoms and a causality assignment will require additional assumptions. For a MB_FSM, causality has already been settled in the process of its construction. The objective here is to explore the human involvement. The objective is *not* to review all existing approaches or to exhaustively review the computing of the fault detection and diagnostics. The presented FSM constructions just survey the essential steps to be taken in the simple case study that is feed forward, linear and time invariant.

### 3.1. Experience based fault signature matrices

This section will outline the construction of an EB_FSM that will be used in the case study in section 4.

Step 1: choose the faults (EB_FSM rows).

The faults of choice should be both (*i*) relevant and (*ii*) present in the history of measurements. In principle, the relevance of a fault resides in the domain of an individual's taste. However, engineering guidelines (ISO, 2016) may establish common sense about typical equipment-, component- (OREDA, 2002) or part level failure modes (Chandler et al., 1991). A Reliability Centered Maintenance (RCM) process may filter out the critical failures, while identifying the faults that may predict them.

The history of measurements will often be collected by non-experimental research which precludes control over the collection of *all* relevant fault states and operating regimes. By conceiving many fault states and operating regimes, the collection of the history of measurement may take too long (=complexity issue analogous to computational complexity). Moreover, faults are often a hidden variable. As already signaled by Tiddens et al. (2023), the history of measurements often delimits aspirations to compute fault detection and isolation.

Step 2: choose the features (EB_FSM columns).

The features of choice should be built from the history of measurements. A data scientist may generate an enormous amount of features from the library of signal features (Lu et al., 2023) while ignoring the choice of the faults. Burnham & Anderson (2002) already argued that even vague knowledge regarding related variables reduces the computational complexity of the model selection while avoiding spurious relations. Engineering guidelines may establish common sense about features (Isermann, 2011) that indicate a fault.

Step 3: select a model

Any regression or classification model may be considered to describe the data, but the shortest description is supposed to be the best one (Occam's razor). However, the shortest description of a data string is fundamentally incomputable (Solomonoff, 1964). Therefore, model selection remains rather arbitrary. Still, a suboptimal approximating model could support a dithering decision maker accepting some risk.

Step 4: explain the model

To explain the selected model, i.e. to identify which features strongly relate to a fault, some arbitrary feature importance test may be chosen. However, feature importance scores do not indicate causality, while a decision maker who does not only seek support in deciding *whether* to act, but also in *how* to act, requires causality. Section 2.4 mentioned that Structural Causal Modelling (SCM) will be used to specify the independence assumptions.

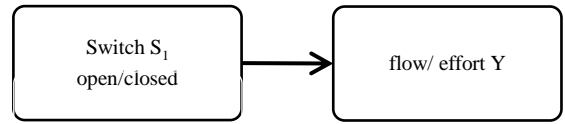

Figure 4. Example of a DAG.

Figure 4 is a directed acyclic graph (DAG) that specifies the causalities in a universe of the variables $(S, Y)$. By Bayesian Network Factorization, the joint probability distribution $Pr(S, Y)$ follows from the DAG in Figure 4:

$$Pr(S,Y) = Pr(S)Pr(Y|S) \qquad (3)$$

Eq. (3) specifies the potentially observable association to identify a causality provided that the DAG is true. For example, the causality $Pr(Y|do(S))$ is identifiable by the potentially observable association $Pr(Y|S)$, provided that Figure 4 is true. The DAG may be highly controversial, but it is explicit at least (Pearl, 2009).

### 3.2. Model based fault signature matrices

This section will outline the construction of a MB_FSM that will be used in the case study.

Step 1: construct an engineering model

A device does not come from some natural phenomenon, it is the result of a deliberate design. Engineers typically compute their designs using the laws of physics. These laws of physics hold under idealized conditions and they should adequately approximate the "real" conditions. These approximations are usually reflected in engineering guidelines that prescribe safety margins. Laws of physics and engineering guidelines are arbitrary in principle as they are occasionally updated, but they generally reflect a very high degree of common sense.

Step 2: choose the faults (MB_FSM rows)

Faults should be phrased in terms of drifts in parameters in the engineering model. If other faults (beyond the parameters in the model) should be detected or isolated, the engineering model needs extension or an additional EB_FSM may be needed.

Step 3: choose the $ARR$s (MB_FSM columns)

From an engineering model of *n* equations the values of *n* variables are computable. As (some of) these variables are measured, less equations are needed which enables the formulation of Analytical Redundancy Relations ($ARR$). An $ARR$ is an equivalence consisting of measurements and parameters from the engineering model. An $ARR$ detects faults that have been defined as parameter drifts, and thus acts as feature or symptom in the FSM.

Step 4: construct the MB_FSM

The faults (MB_FSM rows) have been defined at step 2. The *ARR*s have been defined at step 3, and the fields trivially follow from the presence of the parameters in the *ARR*s. Therefore, the construction of the MB_FSM is autonomously computable from the previous steps.

The *ARR*s are acausal equivalence relations. However, computing the solution of the *ARR*s involves a sequential process where the values of the *ARR*s follow from their variable and parameter values. Similarly, a "real" decision to set a variable or a parameter to a specific value causes the corresponding *ARR*s to change. As a fault in step 2 has been defined as a drift in some *ARR* parameter, this fault causes the *ARR*s to change within the universe of idealized conditions of the engineering (design) model from step 1.

## 4. CASE STUDY

This section will demonstrate fault detection and isolation by constructing an EB_FSM and a MB_FSM in an iconic case study of a linear time invariant system under feed-forward control. This case study involves the RRC circuit in Figure 5.

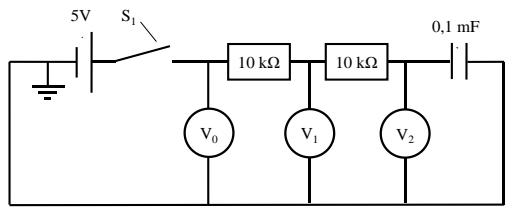

Figure 5. The RRC circuit.

A pulse signal with a period of 20 seconds will trigger the switch $S_1$. The lines in Figure 6 show the computed evolution of the voltages and the dots show the measured evolution of the voltages for a normal (healthy) state of the circuit.

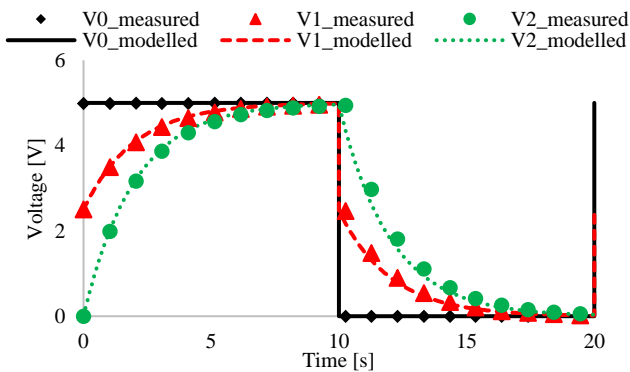

Figure 6. Evolution of the computed and the measured voltages at the healthy state.

Figure 6 confirms that engineers are highly capable of deciding about the "real" behavior of the RRC circuit by computation. Occasionally, the "real" measurements may drift away from the engineering computation which could trigger maintenance. In this case study, two fault treatments have been applied:

1. A decreased resistance $R_0$ that is in between the voltages $V_0, V_1$ in Figure 5.
2. An increased capacitance.

Fault detection and isolation would have been trivial if the resistance and the capacitance were directly observable. It is only due to the experimental setup of this case study that the presence and absence of the faults was certain. Therefore, fault labels in Figure 7 and Figure 8 just followed from a known experimental intervention.

Figure 7 shows that in the faulty state (reduced resistance) the measured voltages respond faster to the switch than predicted by the engineering computation (for the healthy state).

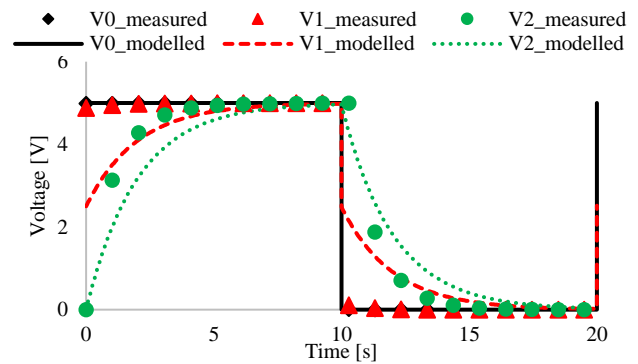

Figure 7. Evolution of the computed and the measured voltages at a decreased resistance R₀.

Figure 8 shows that the measured voltages respond slower to the switch at an increased capacitance than predicted by the engineering computation.

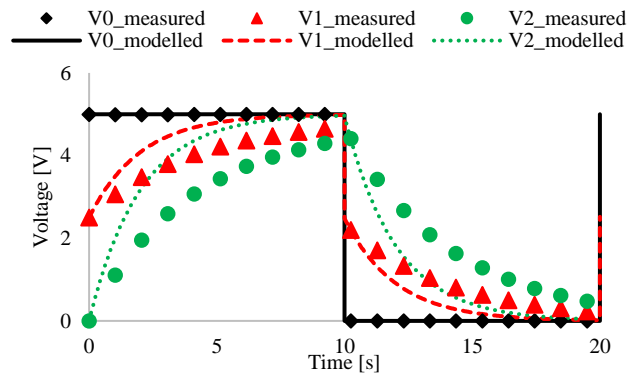

Figure 8. Evolution of the computed and the measured voltages at an increased capacitance.

Note that the operating regime of a pulse signal highly influences Figure 6, Figure 7 and Figure 8 as the RRC circuit is known to operate as a low-pass filter.

Section 4.1 and section 4.2 will explain the construction of an EB_FSM and a MB_FSM respectively. The focus will be on the possible obstructions (section 2.1) in computing fault detection and isolation and not on a quest for the optimal computation.

### 4.1. Application of EB_FSM

Let the faults (EB_FSM rows) be a reduced resistance and an increased capacitance. Let the features (EB_FSM columns) be the *measured* voltages $V_0, V_1, V_2$, the switch position $S_1$, and the time $T$ from Figure 6, Figure 7, and Figure 8. Note that the lines in the three plots are the predictions of an engineering (design) model that should be ignored here.

Let the fields of the EB_FSM be the permutation importance scores of an arbitrary random forest classification. The permutation importance indicates the mean Gini impurity loss of the random forest classification after random resampling of a feature. Note that the EB_FSM fields do not only rely on aforementioned choices, but also on the history of measurements in Figure 6, Figure 7, and Figure 8.

Then, the EB_FSM is given in Table 2.

Table 2: EB_FSM of the case study.

|  | $V_0$ | $V_1$ | $V_2$ | $T$ | $S_1$ |
|---|---|---|---|---|---|
| Resistance $R_0$ ↓ | 0,00 | 0,30 | 0,06 | 0,04 | 0,00 |
| Capacitance ↑ | 0,04 | 0,18 | 0,12 | 0,16 | 0,00 |

Table 2 shows that the voltage $V_1$ entailed unique information about a decreased resistance as random resampling strongly affects the mean Gini impurity loss of the random forest classification. Similarly, the voltages $V_1, V_2$, and the time $T$ entailed unique information about an increased capacitance.

The EB_FSM may be used to reduce the complexity of the model selection as Table 2 implies that the random forest classification could still detect both faults when the switch position $S_1$ is omitted from the history of measurements.

As this paper is not about an improved model selection, details about the arbitrarily selected model will be omitted. It has just been verified that the model of choice correctly predicted all instances in a validation set comprising the same faults that occurred during the same operating regime. So, fault detection and fault isolation (Figure 2) is possible for this specific validation set.

Let the DAG in Figure 9 apply to the EB_FSM (Table 2). This DAG asserts that changes in the resistance $R_0$, in the capacitance $C$, or in the switch $S_1$ cause some hidden flow and effort variables. However, these flow and effort variables are indicated by the voltages $V_0, V_1, V_2$.

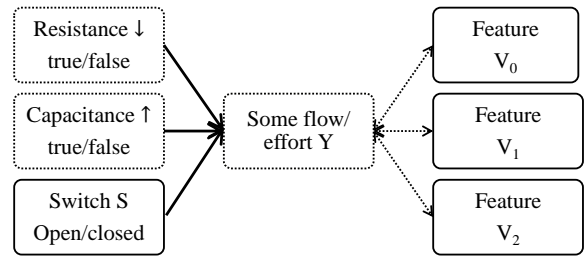

Figure 9. DAG with indicators.

It has been presumed that the switch $S_1$ in the DAG (Figure 9) does not cause the faults and the EB_FSM confirms that the switch $S_1$ neither associates with the faults. Similarly, it has been presumed that the time $T$ does not cause the faults (not in DAG) but the EB_FSM shows that the time $T$ still associates with the faults. Still, section 2.4 already mentioned that observed associations (in the EB_FSM) are not compelling for a DAG. A DAG merely specifies the independence assumptions (omitted arrows) of a specific causal explanation for the EB_FSM.

Section 3.1 mentioned that a decision regarding the fault detection or isolation may be incomputable because it is fundamentally incomputable, it is too complex, or it is subject to philosophical controversy. In this case study, the latter prevailed as the DAG is merely postulated afterwards. Therefore, a compelling causal explanation of the computed fault detection and isolation is lacking. In other words, the causality is philosophically controversial. Common sense reflected in engineering guidelines (section 3.1) may alleviate this controversy. The effects of this controversy are:

- Fault detection and isolation beyond the history of measurements (training set) is risky.
- The applicability of the fault detection and isolation is unknown, i.e. it worked for a specific validation set, but it is unknown whether it will work at an unprecedented operating regime.
- The features (like the time $T$) do not necessarily indicate the magnitude of the fault.

Finally, the fault detection and isolation relied on the arbitrary choice of the classification model, and the feature importance score. Different results might have been obtained had other choices been made.

### 4.2. Application of MB_FSM

In advance of constructing a MB_FSM, an engineering (design) model will be posited. Let the case study be represented by the Hybrid Bond Graph (HBG) in Figure 10.

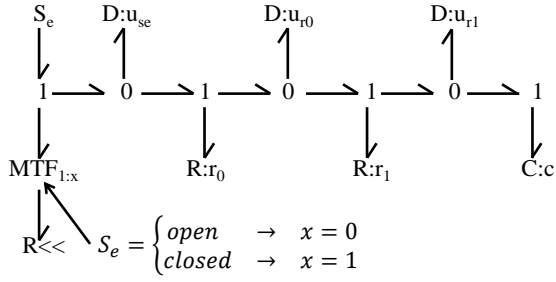

Figure 10. Hybrid Bond Graph of the case study.

The switch has been modelled by a modulated transformer (MTF) as proposed by Borutzky (2012). Figure 10 shows four elements that convert power. As power is the product of an effort variable and a flow variable, the engineering model (Table 3) consists of eight variables and eight constitutive equations that follow from Ohm's Law and Kirchhoff's Law. In this case study, the effort of the source $u_{S_e} = V_0$, and the effort of the resistances $u_{R_0} = V_0 - V_1$, $u_{R_1} = V_1 - V_2$ have been measured which makes three of the equations in Table 3 redundant.

Table 3: Engineering (design) model for case study.

| i | $u_{S_e} = 5 \times x\ ; x \in \{0,1\}$ |
|---|---|
| ii | $0 = u_{R_0} - 10^4 \times i_{R_0}$ |
| iii | $0 = u_{R_1} - 10^4 \times i_{R_1}$ |
| iv | $0 = u_C - 10^4 \times \int i_C(t)dt$ |
| v | $0 = u_{R_0} + u_{R_1} + u_C - u_{S_e}$ |
| vi | $0 = i_{R_0} - i_{R_1}$ |
| vii | $0 = i_{R_0} - i_C$ |
| viii | $0 = i_{R_0} - i_{S_e}$ |

Let's now construct an MB_FSM of the case study using this engineering model. Let the faults (MB_FSM rows) be a drift in the resistance $R_0$ and a drift in the capacitance $C$. As a drift may include an increase as well as a decrease, these fault definitions are more generic than the ones in Figure 7 and Figure 8. Note that the history of measurements (Figure 6, Figure 7 and Figure 8) is not needed for the construction of a MB_FSM.

Let the features (MB_FSM columns) be defined by the ARRs that follow from the measured variables in the engineering model (Borutzky, 2021), (Samantaray et al., 2006).

The $ARR_1$ is given by:

$$0 = \frac{V_0 - V_1}{R_0} - \frac{V_1 - V_2}{R_1} \quad (4)$$

The $ARR_1$ follows from (ii), (iii) and (vi) in Table 3, and the voltages $V_0$, $V_1$, and $V_2$.

The $ARR_2$ is given by:

$$0 = V_0 - V_2 - (V_{0x} - V_{2x}) \times e^{-\frac{(T-x)\times C}{R_0+R_1}} \quad (5)$$

In Eq. (5), $V_{0x}, V_{2x}$ represent the voltages at the time of the last switch transition. The $ARR_2$ follows from (iv) and (v) in Table 3, the evolution in $u_{S_e}$, and the measurements $V_0$, $V_2$, $T$.

Let the fields of the MB_FSM be given as shown in Table 4, revealing an indicator function on the presence of the drifting parameters in the ARRs..

Table 4: MB_FSM of the case study.

|  | $ARR_1$ | $ARR_2$ |
|---|---|---|
| Drift in $R_0$ | 1 | 1 |
| Drift in $C$ | 0 | 1 |

Now, the MB_FSM could be used to evaluate the same validation set as the one used for the EB_FSM. Figure 11 confirms that both ARRs drift away from zero at a decreased resistance as predicted in the MB_FSM (Table 4).

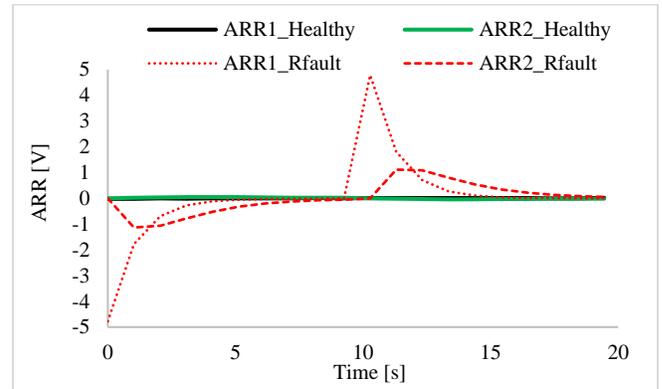

Figure 11. Measured ARRs at a decreased resistance.

Figure 12 confirms that only $ARR_2$ drifts away from zero at an increased capacitance as predicted in the MB_FSM (Table 4). By choosing a threshold value for the ARRs, the fault detection is autonomously computable. Figure 11 and Figure 12 show that the ARR s can only detect faults as the components in the RRC circuit exchange power shortly after a transition of the switch. The applicability of the fault detection and isolation under various switching regimes is straightforwardly assessable without any history of measurements.

Eq. (4) and Eq. (5) specify the value of the $ARR$ at a given magnitude of the drift in $R_0$ or $C$, i.e. a decision regarding the fault identification (i.e. severity of the fault) is partially computable. An autonomously computable fault identification implies a higher maturity in data driven maintenance support (Figure 2) than just isolating the fault. Moreover, the impact of the precision of the measurements is assessable at the stage of design. The precision of the measurements is important to define appropriate threshold values on the $ARR$s.

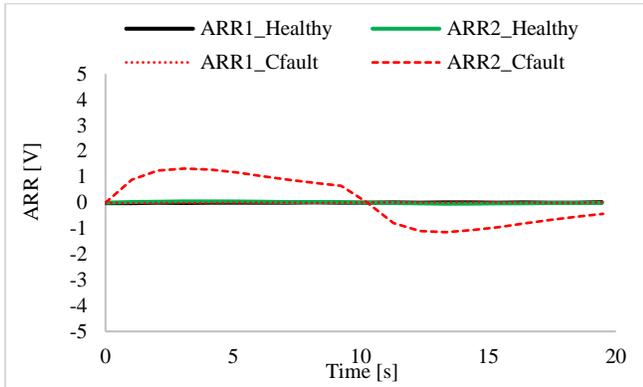

Figure 12. Measured ARRs at an increased capacitance.

Section 2.1 mentioned that a decision regarding the fault detection or isolation may be incomputable because it is fundamentally incomputable, it is too complex, or it is philosophically controversial. In this case study, the latter prevailed as the engineering model is not pertinently true. The fault detection and isolation relied on the applicability of the idealized conditions of the laws of physics that underlie the engineering model. Typically, physical laws are rather robust against changes in these conditions. Still, unmeasured operating conditions may become problematic. For example, large but unrecorded temperature fluctuations may trouble Ohm's Law and consequently the fault detection and isolation (MIL-HDBK-217F, 1991).

If the engineering (design) model were to be true, the MB approach would have resolved the concerns of the EB approach:

- Fault detection and isolation beyond the history of measurements (training set) is decidable. The MB_FSM can even be constructed at the stage of design (without any training set at all).
- The applicability of the fault detection and isolation to work is known. For example, it is known that the fault detection and isolation only works as power is being exchanged.
- The $ARR$s indicate the magnitude of the fault. Therefore, the attainable maturity in data driven maintenance is potentially higher.

Finally, section 3.2 mentioned that the engineering (design) model may just be incapable to detect or isolate a particular fault. As aspirations should meet capabilities, the engineering (design) model may need adjustments for the purpose of data driven maintenance.

## 5. DISCUSSION

This section will reflect on the case study. Section 5.1 will discuss the impact on the computability of "real" decisions, section 5.2 will discuss the impact on the maturity in data driven maintenance, and section 5.3 will discuss some practical implications.

### 5.1. Impact on computing "real" decisions

Section 2.1 mentioned that a decision may be incomputable because it (*i*) is fundamentally incomputable, it (*ii*) is too complex, or it (*iii*) is philosophically controversial.

In this simple case study, the philosophical concerns appeared predominant as the translation between a syntactical computation and a "real" decision required arbitrary human involvement to choose:

- The faults (EB, MB);
- The measurements/ features (EB/MB);
- A classification model (EB/MB);
- A feature importance score (EB);
- A causal explanation (EB);
- An engineering (design) model (MB).

The engineering profession established a high degree of common sense regarding this translation by formulating laws of physics and guidelines. This common sense lacks the solidity of a mathematical proof, and it has been subject to occasional improvement, but it has shown to be effective due to the wide application of engineered devices. Section 2.2 stated that where engineers fail to compute "real" decisions, a human involved maintenance control loop is typically triggered. Still, parts of the maintenance control loop may be computed as shown in the case study. Cases where the computing of "real" decisions is challenging, are also expected to be of high interest to scientists.

In the simple case study, complexity was not an issue. Still, complexity plays a role in other cases. For the EB approach, the inference of a high dimensional model from a large history of measurements may require excessive computing time. Section 3.1 stated that complexity may impede the collection of a history of measurements that includes all relevant system states. Particularly under a non-experimental research construct, the required time is uncontrolled. For the MB approach, the solving of a high dimensional engineering (design) model may similarly bump into complexity concerns.

Fundamental incomputability precluded the selection of a true EB_FSM model (section 3.1). Similarly, the truth of the engineering model (Table 3) was ultimately an incomputable postulate. Fundamental incomputability is also an issue in cases of software faults as there cannot exist a computing device that separates looping software from software that halts in the general case. If this computing device only had to separate software of some fixed number of input symbols, the computation rapidly becomes too complex to solve in time (Rado, 1962).

### 5.2. Impact on maturity

Growth in the data maturity model (Figure 3) coincided with the flow of the consecutive decisions in the maintenance control loop (Figure 2). This paper confirms that the computation of fault detection and isolation should be settled before addressing the computation of decisions further downstream the maintenance control loop. Similarly, maturity growth in data driven maintenance should start with computing fault detection and isolation.

In the specific validation set of the case study, the EB approach and the MB approach were exchangeable in terms of missed and false alarms. Still, a decision maker should not be indifferent towards the approach because (*i*) causality is assigned differently, and (*ii*) the meaning of the features differs. Using the EB approach, causality was assigned afterwards using some arbitrary DAG and the features just described the state of the RRC circuit. Using the MB approach, causality was inherent in the solving of the engineering (design) model and the *ARR*s represented the magnitude of the fault. The latter is part of fault identification (Figure 2) which corresponds with a higher maturity in data driven maintenance.

### 5.3. Practical impact

The case study revealed that the "real" causal implications of some syntactical computation matter for the attainable maturity in data driven maintenance. In the cases study, both the EB and the MB approach appeared to be not entirely compelling for causality. Still, some references to engineering guidelines were given to alleviate potential controversy. Section 3.1 referred to some engineering guidelines for (*i*) the most relevant faults of specific devices and for (*ii*) typical features to detect these faults. Section 3.2 referred to some engineering guidelines to establish common sense regarding the margins between the computed strength and the "real" strength.

For this iconic case study, the construction of a MB_FSM was easy but for a more realistic case study, the construction of a MB_FSM could become complex. Typically, the knowledge of the engineering models is scattered over various agents who may be unwilling to share them. Consequently, much effort may be wasted on reconstructing design models that are in principle already available. Life cycle modelling as proposed in ISO (2014) is a precondition to apply a MB_FSM efficiently in practice.

The EB approach and the MB approach do not compete as one may also consider a hybrid FSM that adds the *ARR*s to an EB_FSM. The EB approach that decides on associated symptoms may be an appreciable resort in the absence of a causal explanation. The MB approach demonstrated the potential of a more mature data driven maintenance under idealized conditions.

### 6. CONCLUSION

This paper argued that some decision problems cannot be solved by any autonomous computation and that maintenance decisions are prone to be computationally challenging. A maturity framework has been proposed that specifies the decisions in a maintenance control loop, and connects these to the aspects of human interpretation, computability and causality. An application of the lowest maturity level to an iconic case study showed that decision makers should not be indifferent to (two) models that provide equal decisions on a validation set in terms of missed and false alarms. Access to a true engineering (design) model allows achieving a higher maturity level in data driven maintenance but it has been observed that a true model cannot be computed from only a history of measurements. Where logic cannot decide, the common sense reflected in engineering guidelines provides a resort at an acceptable risk.

### ACKNOWLEDGEMENTS

This research is part of the European Digital Naval Foundation (EDINAF) Project, a project that has received funding from the European Defence Fund (EDF) under grant agreement 101103273 - EDINAF - EDF-2021-NAVAL-R-2. Funded by the European Union. Views and opinions expressed are however those of the author(s) only and do not necessarily reflect those of the European Union or European Commission (the granting authority). Neither the European Union nor the granting authority can be held responsible for them.